# Secure and Privacy Preserving Proxy Biometrics Identities

Harkeerat Kaur, Rishabh Shukla, Isao Echizen and Pritee Khanna

**Abstract** With large-scale adaption to biometric based applications, security and privacy of biometrics is utmost important especially when operating in unsupervised online mode. This work proposes a novel approach for generating new artificial fingerprints also called '**proxy fingerprints**' that are natural looking, non-invertible, revocable and privacy preserving. These proxy biometrics can be generated from original ones only with the help of a user-specific key. Instead of using the original fingerprint, these proxy templates can be used anywhere with same convenience. The manuscripts walks through an interesting way in which proxy fingerprints of different types can be generated and how they can be combined with use-specific keys to provide revocability and cancelability in case of compromise. Using the proposed approach a proxy dataset is generated from samples belonging to Anguli fingerprint database. Matching experiments were performed on the new set which is 5 times larger than the original, and it was found that their performance is at par with 0% FAR and 0%FRR in the stolen key/ safe key scenarios. Other parameters on revocability and diversity are also analyzed for protection performance.

———————————————

Harkeerat Kaur
Indian Institute of Technology Jammu, India, e-mail: harkeerat.kaur@iitjammu.ac.in

Rishabh Shukla
Indian Institute of Technology Jammu, India, e-mail: 2021rcs2012@iitjammu.ac.in

Isao Echizen
National Institute of Informatics, Tokyo, Japan, e-mail: iechizen@nii.ac.jp

Pritee Khanna
Indian Institute of Information Technology Design and Manufacturing Jabalpur, India, e-mail: pkhanna@iiitdmj.ac.in

# 1 Introduction

Recent advances have witnessed huge growth in biometric enabled online identity verification and access authentication. We can now experience all form of life activities like teaching, learning, shopping, interacting, playing, working in the virtual world, popularly also called as the Metaverse which was once just an imaginary con- cept. We have already begun doing biometric based online monetary transactions, access control, possession of virtual real estate, marriage registry and digital avatars. As one application of the metaverse, why not envision being able to posses a unique proxy biometric (face, eyes, fingerprint, etc) in the cyberspace? A proxy biometric is a virtual agent of an actual user which can act on its behalf and represent him/her in the metaverse with a unique and privacy preserving identity. An identity that can link us from the physical to the digital world, ensure our presence, and at the same time safeguard the revealing of our personal information.

In the present day scenario we are directly converting our original biometric to digital ones. Although, coupled with various protection mechanisms, once your 'digital biometric' template leaves your device, the owner has no information over where and how is its stored, shared or sold for profit, surveillance, or has been victimized to data hacks. As the expected number of metaverse applications will increase, it becomes imperative to question how can we exist online with digital versions of our biometrics and possibility of biometric avatars?

This work proposes concept of '**Proxy Biometrics**' for creating new digital fingerprint biometric templates that will operate on behalf of user's actual biometric. These artificial fingerprints will be generated from original ones coupled with a sec- ond factor key/PIN using deep neural networks (DNNs). Most importantly they will look like real templates so that online applications can extract discriminative infor- mation from them just like original templates. Third parties can process and encrypt proxy biometric in the way they want without user having to be worried about its loss of personal information. In case of a compromise, only the proxy fingerprint will be lost which cannot be inverted to original one. Unlike only ten fingerprints, a user can have multiple proxy fingerprints by coupling different keys.

The organization of this work is as follows. Section 2 discuss the timeline of biometric template protection techniques, followed by the model design of the pro- posed approach in Section 3. The proxy fingerprints are analyzed for performance in Section 4 followed by conclusions in Section 5.

# 2 Biometric Template Protection Timeline

A biometric authentication system is susceptible to various types of attacks where the identity can be compromised at various points between collection, transmission, storage and matching [17]. Biometric data stored in remote databases is a highly vulnerable to hacking, personal data mining, reconstruction attacks, and sharing for covert surveillance and tracking. Templates stored in database once leaked ren-ders the user to a continuous risk of identity thefts and exposure of sensitive information. In order to address this problem various template protection techniques have been proposed which can be majorly categorized as - biometric cryptosystems, cancelable biometrics, and secret sharing techniques [10]. Soutar et. al (1999) pro- posed biometric encryption technique to provide security to stored templates [20]. Fuzzy commitment, fuzzy vaults, homomorphic encryptions and secure sketches were other design techniques proposed under the category of biometric cryptosys- tems [9]. Over the period of time various correlation and reconstruction attacks have been proved on these approaches and the schemes underwent various improvisations to combat them [18]. However, one major drawback is that these schemes heavily rely on security of encryption keys and do not provide the ability to revoke.

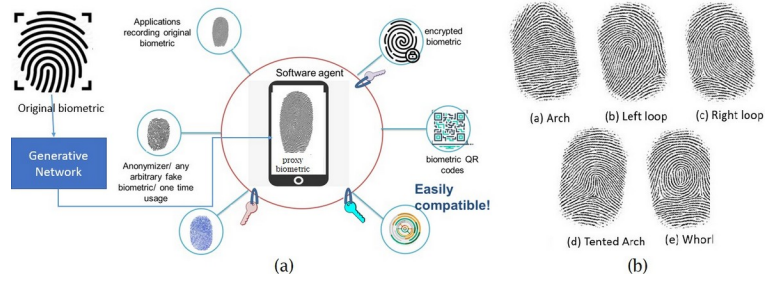

**Fig. 1** Overview original to proxy fingerprints for metaverse applications.

Ratha et. al (2001) proposed the concept of cancelable biometrics. It subjects biometric template to various systematic and non linear distortions combined with a user-specific key in order to generate protected template [17]. The transformed tem- plates are non-invertible and cannot be traced back to the original even if the key and the transformation function is known. The combination of user-specific key provides multi-factor security and easy revocation. One of the most popular transformation technique is BioHashing (2004), which belongs to the category of random projec- tion (RP) based transformations. The data is projected on user-specific orthonormal vectors followed by non-linear quantization to impart non-invertibly to generate a protect code know as BioCode [7]. Multiple random projections, multispace ran- dom projections (2007), dynamic random projections (2010), feature adaptive ran- dom projections (2021), and a number of RP based transformations have been pro- posed along this line to improved security and matching performance [22, 25, 24]. Random Convolution based transformed is another category under which biomet- ric feature is convolved with user specific random kernels. BioConvolving (2010) and curtailed circular convolutions (2014) are two important approaches under this transform [14, 23]. Apart from that a number of techniques have been proposed that combine biometric data with synthetic noise or patters to generate transformed ver- sions. Geometric transformations driven by random noise signals (2006), BioPhasor (2007), Gray-Salting/Bin-Salting (2008), locality sensitive hashing like Indexing- First-One (2017) and Index-of-Max (2018) have been proposed [21, 26, 16, 8].

With advances in deep learning from 2018-19, a number of DNN techniques have been proposed that transform the features simultaneously or separately using one of the transformation techniques described above. Liu et al. (2018) proposed FVR-DLRP for generating secure fingervein templates which used random projection methods to distort the extracted features later trained over Deep Belief Network (DBN) for accurate matching [13]. In 2018, another work by Singh et al. proposes a cancelable knuckle prints which are generated by computing local binary pattern over CNN architecture [19]. Jo and Chang (2018) proposed CNN-based features extracted from face templates into binary code by using a Deep Table-based Hash- ing (DTH) framework [6]. Lee et al. (2021) proposed SoftmaxOut Transformation- Permutation Network (SOTPN) which is a neural version of popular cancelable transform known as Random Permutation Maxout (RPM) transform [11].

However one of the major of all the above mentioned approached is that the trans- formed template has a noise like appearance. These unstructured templates makes matching application specific and effects performance. Some of them would prefer cryptosystem approach, some would like cancelable biometrics, while there may be some who would deploy none due to technical or real time challenges. Literature also proposed various approaches to generate fake biometric using GANS (gener- ative adversarial networks) [15, 3]. However they are useful only for large scale database generation for training purposes only.

The major motivation of this work is to provide a protection layer at user-level which allows generation of artificial fingerprints from original fingerprints combined with a user-specific key. These artificial fingerprints can then act as cyberprox- ies of original and interact with third parties applications for storage and transmis- sion purposes. Fig. 1 shows an illustration which depicts proxy template generated at smartphone interacting with various third parties. The major contributions are:

1. Generation of proxy fingerprint biometrics, that are natural looking, revocable, posses good descriptive characteristics and can only be derived from original fingerprints.
2. Generation of new fingerprint samples of different class from same original by using random projection technique and key.
3. A framework which allows enrollment, authentication, and easy revocation of proxy fingerprints by changing keys in the metaverse.

## 3 Proposed Approach

**Overview** :Based upon their dominant patterns the fingerprints belong to five differ- ent type of classes namely - whorl, arch, tented, left loop, right loop [2], Fig. 1. The proposed is divided into four important steps as shown in Fig. 2. The encoder takes an original fingerprint image as input from the user and extract a latent vector. It is then mixed with some user-specific random key to generated a transformed feature as called as salted latent vector. The salted latent vector is projected on an orthonor-mal basis which is finally passed through a decoder to generate a new fingerprint image that belongs to one of the five classes mentioned above.

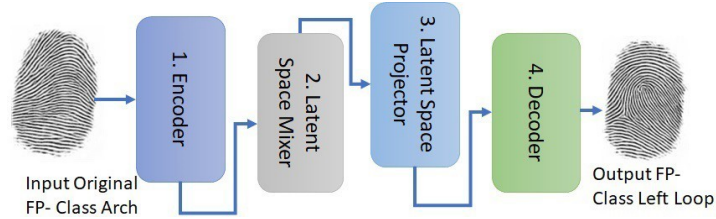

**Fig. 2** Overview of the proposed approach.

**3.1 Design Autoencoder to extract latent spaces**: At the first step we train a CNN based auto-encoder and decoder. The encoder $f_{enc}(x, \theta)$ takes the original finger- print sample $x$ and maps it to latent space representation $z$. The decoder $f_{dec}(z, \varphi)$ reconstructs the fingerprint image $x^{\hat{}}$ from latent representation $z$. Given that the input image sample $x \sim X^{H \times W \times 3}$ and latent vector $z \sim Z$, the image reconstruction loss

$L_{rec}$ for the network is $L_{rec}(f_{enc}, f_{dec}) = E_{x \sim X} [//x - f_{dec}(f_{enc}(x)) //].$

The encoder of a trained autoencoder outputs discriminative latent vector that is often used as input to various other tasks like image denoising, classification, anomaly detection etc. Fig. 3 shows the encoder and decoder architecture. The input image $x \in X^{200 \times 136 \times 3}$ is re-scaled in the range [0,1] before passing. The output taken from the last layer of encoder is flattend to give latent vector $z \in Z^{13600}$. A database consisting of 30,000 synthetic fingerprints spanning across five classes is generated using Anguli software out of which 27000 images were used for training [2].

**3.2 Visualizing and Projecting Latent Spaces**: Once trained for perfect reconstruc- tion, the latent vector $z$ outputted from 'encoders' and their corresponding class and subject labels were collected. Although, the encoded representations are dense and contain discriminative features but are also highly entangled and not well separated. Fig 4(a). shows the the result of conducting principal component analysis over the original latent space. The overlapping distributions are inseparable in terms of their class patterns can be easily observed.

Our model requires to have highly discriminative feature space which can be used to generate new samples of different class. The proposed approach tries to maximize the separations between these overlapping distributions using a technique called Random Projection. By projecting data on orthonormal basis, the pairwise distances of the points before and after projection are not changed thus retaining their statistical properties [4]. The proposed approach utilizes this property to make latent vectors belonging to a specific class follow a certain distribution pattern. The transformation process for the same is outlined as follows.

Step 1:  Step 1. For each class $\ `i\ `$, generate an orthonormal random matrix $M^i$, of dimension 136 $\times$ 136 using the Gram-Schmidt process, where $i \in [1, 5]$ [12].

Step 2:  The latent vector $zp$ is reshaped into a 2D such that $zp \in Z^{100 \times 136}$.

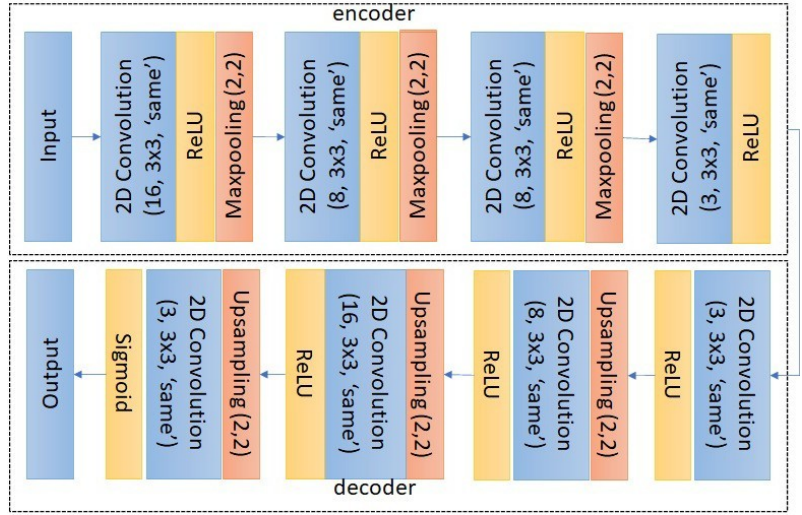

**Fig. 3** Encoder and Decoder Architecture.

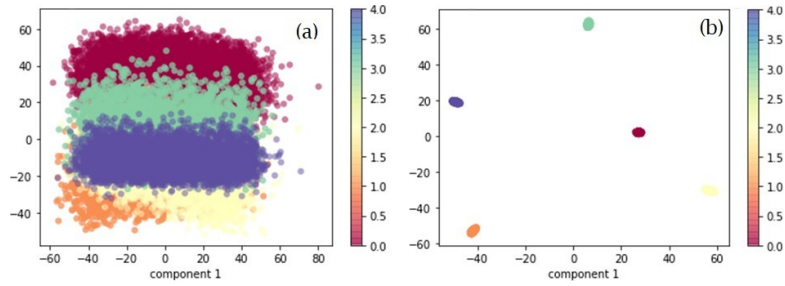

**Fig. 4** Visualizing PCA over latent space (a) Original Space (b) After Projection.

Step 3:  Project $zp$ belonging onto matrix $M_i$ as $P\,j_i = z_i M_i$.
Step 4:  Reshape $P\,j_i$ again in one dimension $P\,j_i \in Zp^{13600}$.

PCA is again performed over the projected latent spaces and results are visualized Fig. 4(b). It can be seen that the projected spaces of latent vectors now follows distribution that is well separated.

**3.2 Preparing Projected Vectors for Decoder** This section aims to propose an approach which trains a decoder model on features belonging to different classes but projected on random matrix $M^i$ to generate new samples belonging of class $i$.

As mentioned, the original database $D$ comprises 30000 fingerprint images labelled to 7500 subjects equally distributed over five classes. With 6000 images per class and with 4 images per subject, the images with subjects numbered 1 to 1500 belong to class 1, 1501 to 3000 to class 2, and so on till 6001 to 7500 to class 5. We take *ith* sample of each class and assign it to new subset $DM_j$ where $j = mod(i, 6)$.

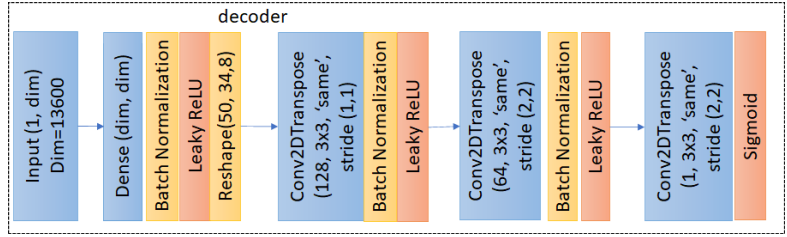

**Fig. 5** Final Decoder Architecture

Thus we have five new subsets of mixed samples named as $Dm_1, Dm_2,...,Dm_5$. The data in subset $Dm_i$ is projected on random matrix $M_i$ to form a projected subset $P_{j_i}$. The decoder takes input the latent vector $P_{j_i} = Zp^{13600}$, where input $P_{j_i}$ consisting of salted latent vectors belonging to any class but projected over matrix and targeted output $x_{dec} \sim X^{200 \times 136 \times 1}$ is a fingerprint image belonging to set $D_i$. The decoder model architecture is depicted in Fig. 5. Since the images are in range [0,1] the final activation is sigmoid. The model is optimized using Adam optimizer to minimize MSSIM (mean structure similarity index) loss between targeted and predicted output images, $L_{dec} = minimize(1 - MSSIM(true, predicted))$.

**3.3 Generating new fingerprint images** Once the encoder and decoder models are trained, it can be used to generate new fingerprint samples.

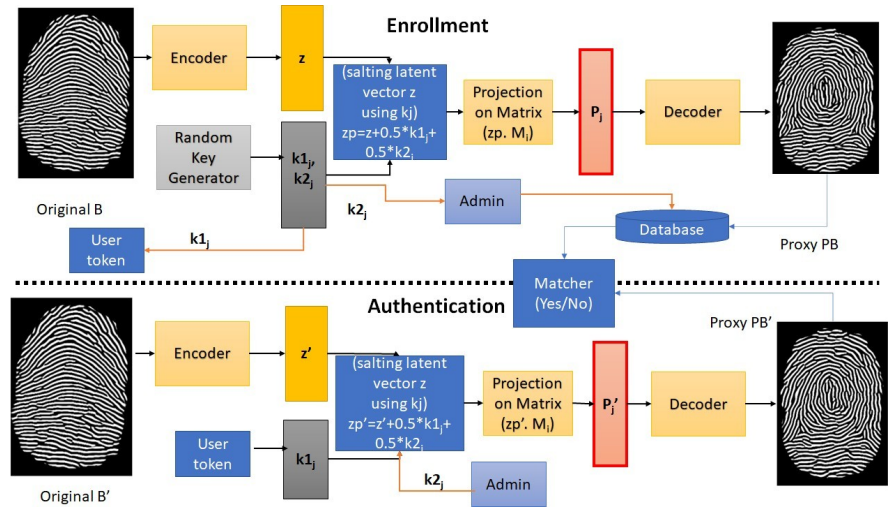

**Fig. 6** Enrollment and Authentication process using proxy biometrics.

This section explains how an existing individual logs on to the proposed architecture to generate a proxy biometric and enroll/register it. Also, how it can generate the probe version of the proxy biometric to gain access into the system.

**Enrollment:** The process is illustrated in Fig. 6 and explained step wise below.

Step 1:  At the time of enrollment a new user inputs its original biometric $B$ to the encoder which extracts its latent vector $z$.

Step 2:  A random key generation unit generates key $k$ with distribution $N$ ($\mu = 0$, $\sigma = 0.5$).

Step 3:  The keys are added to latent vector $z$ and the result salted vector $zp$, generated as $zp = z + k$

Step 4:  The target class is selected, let say class $C_i$. The salted vector $zp$ is projected on the orthonormal random matrix $M_i$ for the class $Ci$ as $P j = zp.M_j$

Step 5:  The Projected vector $P j$ is decoded to output a new proxy biometric $PB$ as

$$PB = decoded(P j)$$

At the end of this process this new proxy biometric $PB$ is recorded as reference template and stored for further matching. The user-specific random key $k$ is divided into two shares $k1$ and $k2$, such that $K = 0.5 * k1 + 0.5 * k2$. The first key share $k1$ and class $C_i$ is provided to the user in tokenized manner. The second key share $k2$ is maintained at the generation site by the system administrator with respect to an identification number provided to the user.

**Authentication:** The steps for authentication are:

Step 1:  The user presents its original probe biometric sample $B^{'}$ and tokenized key $k1$ and target class $C_i$.

Step 2:  The latent vector $z^{'}$ is extracted, second key $k2$ is fetched and salted vector $zp^{'}$ is generated as $zp^{'} = z^{'} + 0.5 * k1 + 0.5 * k2$

Step 3:  The salted vector is projected on targeted class $C_i$ as $P j^{'} = zp^{'}.M_i$

Step 4:  Finally, the projected vector is decoded to form the new proxy probe biometric $PB^{'}$.

Instead of matching the original ones, the proxy samples $PB$ and $PB^{'}$ are matched to grant and deny access. This offers great security from revealing your actual fin-gerprint yet at the same time ensuring your presence. Due to non-linearity imparted by various model, even if the proxy biometric and key are revealed to the attacker, he/she cannot use it to recover the original ones.

## 4 Experimental Results and Analysis

Proposed approach is evaluated on various performance and protection parameters to verify the effectiveness of the generated artificial/proxy fingerprint templates.

**4.1 Matching Performance**: As mentioned this work uses synthetically generated fingerprint form Anguli software since most of the publicly available fingerprint dataset as quite small in size for training DNNs. Given a synthetic database of total 7500 subjects, 25 subjects are selected randomly from each class, thus giving a total database of 125 subjects with 4 samples per subject. Let this data subset be called as "*original db*'. The new biometrics are generated by proposed combination of original and some random keys after passing through various layers of convolutions neural networks. ThE matching performance is evaluated keeping in account of the generating parameters under two scenarios, namely - *worst case or stolen token sce- nario* and *best case or safe token scenario*. It is expected that these proxy biometric must exhibit same uniqueness as well as intra and inter-user discriminating charac- teristics. To match the fingerprint, one of the most common brute force based key- point matching algorithm is used [1]. The brute force algorithm matches keypoints of one fingerprint with another in one to all mode. Apart from that since fingerprint are highly structured in terms of constituting of ridges and bifurcations, Structure Similarity Index (SSIM) has been proven to be an effective metric to record match- ing similarity between two images [5]. SSIM is computed locally over small image patches (here of size $10 \times 10$) and overall mean is computed as Mean Structural Similarity Index (MSSIM), $MSSIM(X,Y) = \frac{1}{M} \sum_{j=1}^{M} SSIM(x_j, y_j)$. Usually the in- dex values are normalized to range between [0 to 1], where 1 indicates perfect sim- ilarity and value 0 otherwise.

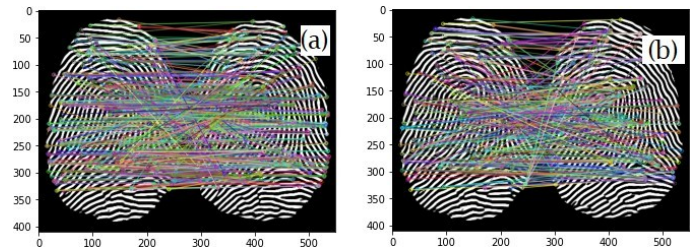

**Fig. 7** Matching using keypoint brute force (a) Intra-user (b) Inter user

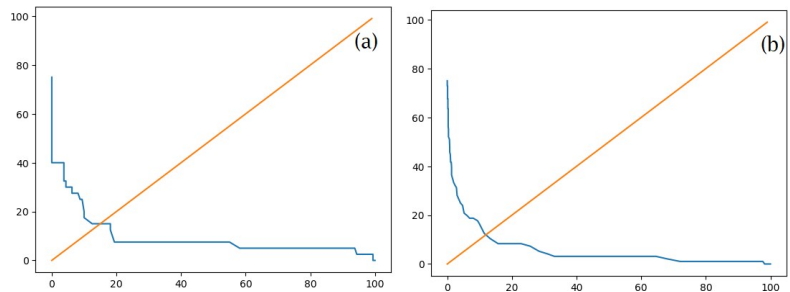

**Fig. 8** ROC curves using brute force technique (a) Original Database (b) Proxy Database in Best Case

**Table 1** Matching Performance in best and worst case scenario.

| Scenario | Database | Brute Force | MSSIM + Brute |
|----------|----------|-------------|---------------|
| Force Original | *original_db* | 18.65 % EER | 0% EER |
| Best Case | *proxy_db* | 21.35 % EER | 0% EER |
| Worst Case | *proxy_db* | 25.35 % EER | 0% EER |

**a) Best Case / Safe Key Scenario**: In this scenario, each user $j$ in *original_db* is assigned a distinct user-specific and system key $k_j =(k1_j, k2_j)$. Five new samples are generated by projecting the salted vector $zp_j$ on matrix $M_1, M_2,...,M_5$, where each sample belongs to particular class. This results in a new dataset called '*proxy db*' having 625 subjects, with 4 samples per subject. This is exactly a five fold increase in population. The proxy samples of each subject $PB_m^x$ and $PB_n^y$ are matched to output a score, where $1 \le m, n \le 625$ (no. of subjects) and $1 \le x, y \le 4$ (samples per subject). If $m = n$, $PB_m^x$ and $PB_n^y$ are expected to match and result in genuine score, otherwise imposter score. Matching scores are computed using brute force and patch-wise MSSIM index between two images as discussed above. Figure 7 illustrates brute force keypoint matching between two proxy fingerprints belonging to same user and different users. Matching performance on '*proxy db*' and similar size subset of *original db* is computed using brute force, and a combination of brute force and MSSIM values. The results in terms of Equal Error Rates are reported in Table 1, Fig. 8. The matching scores for MSSIM+brute force were recorded and the genuine and impostor score distribution over *proxy db* is depicted in Fig.9(a). A clear separation is observed. Thus indicating that the proxy templates preserve good discriminating characteristics and perform desirably with 0% False Accept Rate and 0% False Reject Rate.

**b) Worst Case / Stolen Key Scenario** The scenario consider the possibility that the user key is stolen and is used by the impostor to gain access to the system. Under this scenario the impostor generates proxy biometric using its own/others biometric and stolen key. It is expected that this attacker generated proxy biometric should not match with the legitimate key holder generated proxy biometric or with templates of other enrolled legitimate entities. To simulate this scenario, all the users are assigned same user specific key. For each subject $j$, the extracted latent vectors are salted, projected, and decoded to form a new dataset called *proxy* $^\star$ *db*. Similarly to above, this dataset has 625 subjects. Samples $PB_m^{\star x}$ and $PB_n^{\star y}$, where $1 \le m, n \le 625$, $1 \le x, y \le 4$ are matched and genuine and imposter scores are recorded (refer Table 1). The score distribution for MSSIM+brute force is depicted in Fig. 9(b) and once again a distinct separation can be observed indicating 0% EER even in the worst case scenario.

**4.2 Revocability and Diversity**: This section overviews the revocability property of the proposed approach by noting the effect of changing user specific key and pro- jection matrices on new sample generation. 6pt

**a). Changing one or both keys:** The latent feature $z$ is extracted from original fingerprint $B$. Then it is mixed with a user specific key $K$ consisting of two parts $k1$ and $k2$ to form basis for new sample generation. This section analyzes the effect of changing one or both key parts on new sample generation while keeping the projec- tion matrix same. For each subject in *original* having 125 subject two key sets are generated $ka=(k1a, k2a)$ and $kb=(k1b, k2b)$ and projected each class matrix $M_i$ two create two sample sets *proxy db key*1 and *proxy db key*2 having 625 subjects and 4 samples each. Then corresponding samples from *proxy_db_key*1 and *proxy db key*2 are matched by computing MSSIM between them. Figure 10 illustrates an original fingerprint sample B and its proxies belonging to class 3 and 4 obtained using key *ka* and *kb* and the MSSIM between them. Similarly, the MSSIM between 2500 sam- ples of two proxy set is computed with mean value 0.49383. Thus indicating same biometric samples when salted with different keys and projected over same matrix show significant diversity.

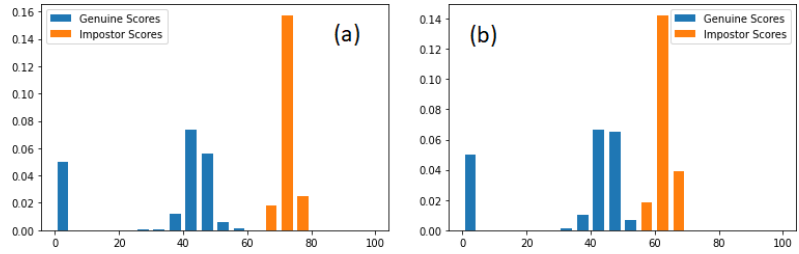

**Fig. 9** Genuine Imposter score distribution (a) Best Case Scenario and (b) Worst Case Scenario.

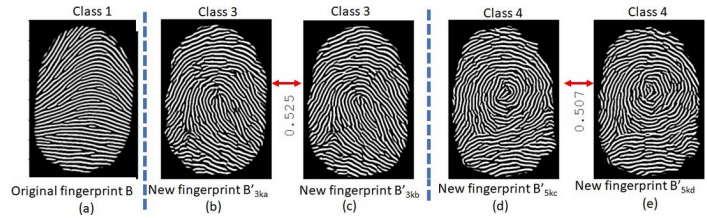

**Fig. 10** Visualizing the effect of changing both keys.

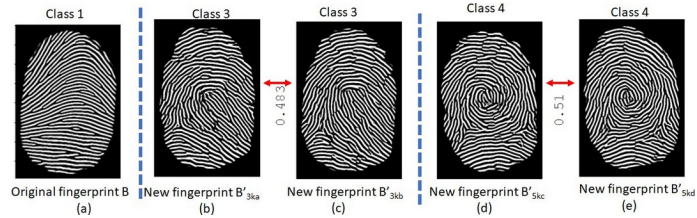

**Fig. 11** Visualizing the effect of changing only one of the key.

On the same line figure Fig. 11 shows the effect by only changing the user key part ($k1a$) and ($k1b$) keeping the second key part same for classes 3 and 5. The generated samples and the MSSIM values between them show that new samples can also be generated by only changing one part of the key. Thus in case of any comprise the one/both keys can be changed and combined with the latent feature to generate new proxy biometric samples as required. The average MSSIM between proxy dataset generated this way is reported to be 0.493738. In both cases average MSSIM value is around 0.5 thus indicating significant in change in overall structure with change in keys $ka$ and $kb$.

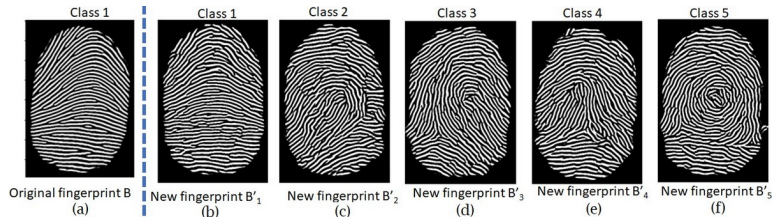

**Fig. 12** Visualizing the effect of changing projection matrix

**b) Changing the projection matrix**: This section analyzes of the effect of changing projection matrix $M_i$ keeping the salted vector same. Given an original fingerprint image $B$ of any class 1, its latent vector $z$ is extracted and salted with a key to get

$zp$. New samples $B'_i$ each class $C_i$ is generated by decoding projection of $zp$ on the respective class matrix $M_i$ as:

$B'_i = decoded((zp).M_i)$ The effect of using different projection on matrix $M_i$ on same $zp$ is depicted in Fig. 12. Figure 12 (a) shows an original biometric sample belonging to class 1 and 12 (b-f) shows proxy samples belonging to class 1 to 5. For each subject in original database five new samples are generated by keeping the "subject" key fixed and changing projection matrix $M_i$. MSSIM is computed between generated samples of different classes having same underlying vector and the average value is reported to be 0.49263. Thus depicting the trained model is capablility of generating new samples having patterns belonging to the projected class $c_i$, when same latent $zp$ is projected on different matrices $M_i$, $i$ =1 to 5.

**4.3 Non-invertibility**: Non-invertibility implies that an attacker should not be able to recover the original biometric from its transformed version even if the key and transformation functions are know. Following that the proposed architecture uses DNNS for the transformation tasks, non-invertibility is implicit here even if the user-specific key and projection matrices are known to the attacker.

## 5 Conclusions and Future Directions

The proposed work provides proof of concept for producing proxy fingerprint tem- plates from the original ones combined with user-specific key. Samples across vari- ous classes can be easily generated by projection along specific orthonormal random matrices for a particular class. With a five fold increase in the population size, the effect of performance on proxy samples under stolen token and safe token scenario were analysed and reported to be satisfactory.

The future work aims at using more realistic dataset for proxy generation as well as testing for other different fingerprint matching schemes. Further increase in population size and improvement in generation model is expect to open new avenues towards secure and privacy preserving biometric authentication in the upcoming meta-world. Overall this work gives a good working proof of how AI will be able to actually manifest natures' ability to generate new biometric samples using DNNs.

**Acknowledgements** This work was partially supported by JSPS Bilateral Program (JSPS-DST) JSPSBP120217719, JSPS KAKENHI Grants JP16H06302, JP18H04120, JP20K23355, JP21H04907, and JP21K18023, and by JST CREST Grants JPMJCR18A6 and JPMJCR20D3, Japan.

## References

[1] (????) Fingerprint detection using opencv 3. https://hub.packtpub.com/fingerprint-detection-using-opencv/

[2] Ansari AH (2011) Generation and storage of large synthetic fingerprint database. ME Thesis, Jul

[3] Bamoriya P, Siddhad G, Kaur H, Khanna P, Ojha A (2022) Dsb-gan: Genera- tion of deep learning based synthetic biometric data. Displays 74:102,267

[4] Dasgupta S, Gupta A (2003) An elementary proof of a theorem of johnson and lindenstrauss. Random Structures & Algorithms 22(1):60–65

[5] Jain AK, Cao K (2015) Fingerprint image analysis: role of orientation patch and ridge structure dictionaries. Geometry driven statistics 121(288):124

[6] Jang YK, Cho NI (2019) Deep face image retrieval for cancelable biomet- ric authentication. In: 2019 16th IEEE International Conference on Advanced Video and Signal Based Surveillance (AVSS), IEEE, pp 1–8

[7] Jin ATB, Ling DNC, Goh A (2004) Biohashing: two factor authentication featuring fingerprint data and tokenised random number. Pattern recognition 37(11):2245–2255

[8] Jin Z, Hwang JY, Lai YL, Kim S, Teoh ABJ (2017) Ranking-based locality sensitive hashing-enabled cancelable biometrics: Index-of-max hashing. IEEE Transactions on Information Forensics and Security 13(2):393–407

[9] Juels A, Wattenberg M (1999) A fuzzy commitment scheme. In: Proceedings of the 6th ACM conference on Computer and communications security, pp 28–36

[10] Kaur H, Khanna P (2016) Biometric template protection using cancelable bio- metrics and visual cryptography techniques. Multimedia Tools and Applica- tions 75(23):16,333–16,361

[11] Lee H, Low CY, Teoh ABJ (2021) Softmaxout transformation-permutation network for facial template protection. In: 2020 25th International Conference on Pattern Recognition (ICPR), IEEE, pp 7558–7565

[12] Leon SJ, Bjo¨rck A˚, Gander W (2013) Gram-schmidt orthogonalization: 100 years and more. Numerical Linear Algebra with Applications 20(3):492–532

[13] Liu Y, Ling J, Liu Z, Shen J, Gao C (2018) Finger vein secure biometric tem- plate generation based on deep learning. Soft Computing 22(7):2257–2265

[14] Maiorana E, Campisi P, Neri A (2011) Bioconvolving: Cancelable templates for a multi-biometrics signature recognition system. In: 2011 IEEE International Systems Conference, IEEE, pp 495–500

[15] Minaee S, Abdolrashidi A (2018) Finger-gan: Generating realistic fingerprint images using connectivity imposed gan. arXiv preprint arXiv:181210482

[16] Ratha N, Connell J, Bolle RM, Chikkerur S (2006) Cancelable biometrics: A case study in fingerprints. In: 18th International Conference on Pattern Recog- nition (ICPR'06), IEEE, vol 4, pp 370–373

[17] Ratha NK, Connell JH, Bolle RM (2001) Enhancing security and privacy in biometrics-based authentication systems. IBM systems Journal 40(3):614–634

[18] Scheirer WJ, Boult TE (2007) Cracking fuzzy vaults and biometric encryption. In: 2007 Biometrics Symposium, IEEE, pp 1–6

[19] Singh A, Hasmukh Patel S, Nigam A (2018) Cancelable knuckle template generation based on lbp-cnn. In: Proceedings of the European Conference on Computer Vision (ECCV) Workshops, pp 0–0

[20] Soutar C, Roberge D, Stoianov A, Gilroy R, Kumar BV (1999) Biometric en- cryption. In: ICSA guide to Cryptography, vol 22, McGraw-Hill New York, p 649

[21] Teoh AB, Ngo DC (2006) Biophasor: Token supplemented cancellable bio- metrics. In: 2006 9th international conference on control, automation, robotics and vision, IEEE, pp 1–5

[22] Teoh ABJ, Yuang CT (2007) Cancelable biometrics realization with multi-space random projections. IEEE Transactions on Systems, Man, and Cyber- netics, Part B (Cybernetics) 37(5):1096–1106

[23] Wang S, Hu J (2014) Design of alignment-free cancelable fingerprint templates via curtailed circular convolution. Pattern Recognition 47(3):1321–1329

[24] Yang B, Hartung D, Simoens K, Busch C (2010) Dynamic random projection for biometric template protection. In: 2010 Fourth IEEE international confer- ence on biometrics: theory, applications and systems (BTAS), IEEE, pp 1–7

[25] Yang W, Wang S, Shahzad M, Zhou W (2021) A cancelable biometric au- thentication system based on feature-adaptive random projection. Journal of Information Security and Applications 58:102,704

[26] Zuo J, Ratha NK, Connell JH (2008) Cancelable iris biometric. In: 2008 19th International conference on pattern recognition, IEEE, pp 1–4